\documentclass[lettersize,journal]{IEEEtran}
\usepackage[utf8]{inputenc}
\usepackage{amsmath,amsfonts}
\usepackage{hyperref}
\usepackage{amssymb}
\usepackage{algorithm}
\usepackage{bbding}
\usepackage{algpseudocode}
\usepackage{color}
\usepackage{multirow}
\usepackage{array}
\usepackage{caption}
\usepackage[caption=false,font=normalsize,labelfont=sf,textfont=sf]{subfig}
\usepackage{textcomp}
\usepackage{stfloats}
\usepackage{url}
\usepackage{verbatim}
\usepackage{graphicx}
\usepackage{cite}
\begin{document}

\title{3DCTN: 3D Convolution-Transformer Network for Point Cloud Classification}

\author{
Dening Lu,
Qian Xie,
Linlin Xu,~\IEEEmembership{Member,~IEEE,}
Jonathan Li,~\IEEEmembership{Senior Member,~IEEE}

\thanks{Corresponding authors: Linlin Xu; Jonathan Li.}
\thanks{Dening Lu, Linlin Xu are with the Department of Systems Design Engineering, University of Waterloo, Waterloo, ON N2L 3G1,  Canada (e-mail: d62lu@uwaterloo.ca; linlinxu618@gmail.com).}
\thanks{Jonathan Li is with the Department of Geography and  Environmental Management, University of Waterloo, Waterloo, ON N2L 3G1,  Canada (e-mail: junli@uwaterloo.ca).}
\thanks{Qian Xie is with the the Department of Computer Science, University of Oxford, Oxford OX1 3QD, U.K. (e-mail: qian.xie@cs.ox.ac.uk).}
}



\maketitle

\begin{abstract}
Although accurate and fast point cloud classification is a fundamental task in 3D applications, it is difficult to achieve this purpose due to the irregularity and disorder of point clouds that make it challenging to achieve effective and efficient global discriminative feature learning. Lately, 3D Transformers have been adopted to improve point cloud processing. Nevertheless, massive Transformer layers tend to incur huge computational and memory costs. This paper presents a novel hierarchical framework that incorporates convolution with Transformer for point cloud classification, named 3D Convolution-Transformer Network (3DCTN), to combine the strong and efficient local feature learning ability of convolution with the remarkable global context modeling capability of Transformer.  Our method has two main modules operating on the downsampling point sets, and each module consists of a multi-scale local feature aggregating (LFA) block and a global feature learning (GFL) block, which are implemented by using Graph Convolution and Transformer respectively.
We also conduct a detailed investigation on a series of Transformer variants to explore better performance for our network. Various experiments on ModelNet40 demonstrate that our method achieves state-of-the-art classification performance, in terms of both accuracy and efficiency. \textit{Our code will be publicly available.}
\end{abstract}


\section{Introduction}
\label{sec:introduction}

With the universal application of sensors that are able to obtain geometric information of 3D scenes, such as 3D laser scanners and RGB-D cameras, 3D point cloud classification, as a basic 3D vision task, has become more and more important for many graphic and vision applications. 3D point cloud data has a great ability of geometrical shape expression, thanks to its simple yet flexible data structure. In recent years, 3D point cloud classification has been applied widely in many important fields, such as urban construction, autonomous driving, robotics, engineering survey and mapping. 
Classification task is highly dependent on global features of the target point cloud. Compared with 2D images, point clouds have more complicated structures, distributing in 3D space in an irregular and disordered manner. So it is still challenging to design deep learning networks to achieve effective and efficient global feature extraction from them.

To tackle the above challenges, a number of deep learning-based approaches on 3D point clouds have been proposed.
Many existing works \cite{wu20153d, maturana2015voxnet, riegler2017octnet, atzmon2018point, le2018pointgrid} focus on projecting the 3D point cloud to 2D parameter planes by multi-view projection, or designing spatial discrete convolutions by introducing 3D space voxelization. Despite the great success in point cloud processing, such methods fail to leverage the sparsity of spatial point clouds, and massive projection operations tend to cause high computation cost and memory consumption.
Different from above approaches, Qi et al. \cite{qi2017pointnet} proposed PointNet to achieve the point cloud feature learning in a point-wise manner. PointNet consists of several core modules: rigid transformations (T-Net), shared Multi-Layer Perceptrons (MLPs) and maxpooling, which ensure the network invariant to point permutation and shape rotation. After that, several variants \cite{qi2017pointnet++, paigwar2019attentional, qi2018frustum} have been proposed to improve the performance of PointNet by introducing local feature extraction. To utilize the strong local modeling ability of convolutional neural networks (CNNs), many meaningful works, such as PointCNN \cite{li2018pointcnn}, PointConv \cite{wu2019pointconv}, and DGCNN \cite{wang2019dynamic}, were proposed to define 3D convolutional kernels or Graph Convolution to achieve point cloud processing and analysis.

Due to the impressive progress in Natural Language Processing (NLP) and computer vision, Transformer has been proven to have a remarkable ability of global feature learning, and thus has been applied to various point cloud processing tasks, such as object classification, semantic scene segmentation, and object part segmentation \cite{guo2021pct, zhao2021point, han2021point,gao2022lft}.
The core component of Transformer is self-attention mechanism, which first computes the similarities between any two embedded words, and then utilizes the corresponding similarities to compute the weighted sum of all words, as the new output.
By this way, each output word is able to establish connections with all input words, which is the reason why Transformer is good at learning global feature. Therefore, current 3D point cloud Transformers tend to employ Transformer to replace all convolution operations in networks for better feature expression.

Despite the success of 3D point cloud Transformers, the efficiency of the Transformer network is still below similarly sized CNN counterparts because of massive linear transformation layers in Transformers. By introducing convolution to the ViT \cite{dosovitskiy2020image} structure, CvT \cite{wu2021cvt} achieves the better performance and robustness, while concurrently maintaining a high degree of computational and memory efficiency. Therefore, in this work, we hypothesize that combining the strong and efficient local modeling ability of CNNs with the remarkable global feature learning ability of Transformers may improve both the accuracy and efficiency for 3D point cloud classification. 

Therefore, we develop a new architecture for point cloud classification, called 3D Convolution-Transformer Network (3DCTN), to incorporate convolution into Transformer, making it inherently efficient and achieve competitive results with state-of-the-art classification methods. Specifically, to avoid computational redundancy, our framework is designed as a hierarchical structure, which has two main modules both operating on the downsampling point sets. And each module consists of two blocks: multi-scale local feature aggregating and global feature learning, which are achieved by Graph Convolution and Transformer respectively.

We evaluate the accuracy and efficiency of our classification network on the public dataset, ModelNet40 \cite{wu20153d}. Extensive results show that our method achieves state-of-the-art classification performance. Additionally, we conduct a detailed investigation and analysis on a series of Transformer variants for better performance, and conclude that the offset-attention mechanism and subtraction-form vector attention operator outperform the other variants for our framework.

In summary, the main contributions of our work are as follows:
\begin{itemize}
    \item We design a highly expressive module combining Transformer and convolution, to learn local and global features effectively for point cloud classification;
    \item Based on such modules, we propose a multi-scale hierarchical framework, which is suitable for the global feature expression of unstructured point clouds;
    \item We conduct a detail investigation and analysis of a series of Transformer variants for better performance;
    \item Extensive experiments on ModelNet40 show that our method achieves state-of-the-art performance in classification accuracy and efficiency.
\end{itemize}

\section{Related Work}
\label{sec:relatedwork}

\begin{figure*}[htbp]
  \centering
  \includegraphics[width=0.9\linewidth]{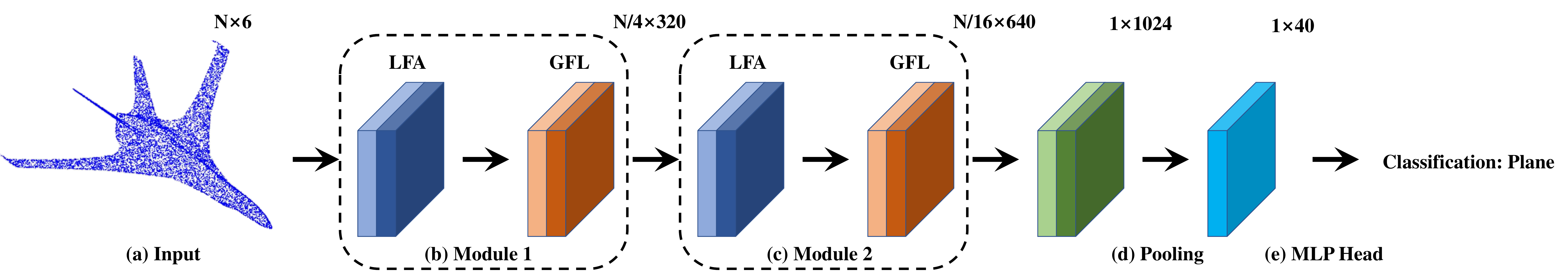}
  \caption{Hierarchical structure of 3DCTN. It mainly consists of two modules, and each of them has a Graph Convolution-based Local Feature Aggregating (LFA) block and a Transformer-based Global Feature Learning (GFL) block.
  \label{fig:overview}}
\end{figure*}

\subsection{3D Point Cloud Classification.}
\textbf{Volume-based Methods.}
Similar as 2D image processing, VoxNet \cite{maturana2015voxnet} introduced 3D voxelization method to point cloud data processing, which is to quantize unstructured point clouds to regular volumetric grid forms, and then the 3D convolutions can be directly applied to point clouds for feature learning.
Such methods fail to leverage the sparsity of spatial point clouds because of rasterization, and it is challenging to construct high-resolution voxelization models due to huge computation and memory costs.
To address these issues, OctNet \cite{riegler2017octnet} proposed an unbalanced grid-octree structure, which allows higher resolution ($256\times256\times256$) input than VoxNet. And \cite{choy20194d} used sparse convolution method, only performing convolution operations at occupied voxels to reduce memory and computation. Despite the great progress volume-based methods made, there still exists the loss of geometric information due to the transformation from irregular point clouds to regular 3D voxels.

\textbf{Projection-based Methods.}
Projection-based methods are also closely related to 2D image processing. MVCNN \cite{su2015multi}, as the pioneer of projection-based methods, projected 3D point clouds into multiple views, where the features of each view are extracted by 2D CNNs, and then aggregated these features through maxpooling. To improve the robustness and accuracy of the view feature aggregation, several variants have been proposed. View-GCN \cite{wei2020view} utilized the Graph Convolution Network (GCN) to establish the relationship between different projection views.
\cite{yu20213d} pointed out the limitations of view-based pooling, and proposed a patch-level pooling method by formulating the view-based 3D classification into a set-to-set matching problem.
However, projection-based methods may incur the loss of geometric information during the projection process, and massive projection views tend to cause high computation cost and memory consumption. Additionally, the number and position of projection views are critical to the classification performance, and it is still challenging to chose projection views adaptively for the underlying geometric structure modeling.

\textbf{Point-based Methods.}
Taking the 3D coordinates or/and normal features as input, point-based methods deal with the unstructured point clouds directly. The early work, PointNet \cite{qi2017pointnet}, was introduced by Qi et. al, where a deep learning network with MLPs and maxpooling was designed to achieve feature learning. After that, to aggregate local features, PointNet++ \cite{qi2017pointnet++} applied PointNet in a hierarchical manner and used query ball grouping to construct local neighborhoods, which proves that the hierarchical structure is effective for point cloud feature learning. To further leverage the local information of point clouds, \cite{li2018pointcnn, wu2019pointconv} introduced 3D convolution kernels to extract local features, instead of shared MLPs. Due to the disorder of neighbor points, PointCNN \cite{li2018pointcnn} introduced $x$-transformation to rearrange the points into a latent and potentially canonical order, followed by using typical convolution to extract local features from point clouds.

Another kind of point-based methods is Graph-CNN, which establishes connections between local points with a graph, and then models the local geometric information. DGCNN \cite{wang2019dynamic} constructed dynamic neighborhoods by features extracted from the former layer, and performed Graph Convolution on neighborhoods. Unlike DGCNN, ECC \cite{simonovsky2017dynamic} defined dynamic convolution-like filters based on edge labels, which also achieves satisfactory results on various datasets. 3DGCN \cite{lin2020convolution} introduced shift and scale-invariance properties to the deep learning networks, and defined learnable kernels with a graph max-pooling mechanism. DeepGCNs \cite{li2019deepgcns} applied residual/dense connections and dilated convolution to GCN frameworks, to train very deep GCNs, which proves the positive effect of depth in GCNs. All these methods show that the Graph Convolution is good at local feature information aggregating, but nearly none of them is designed to model long-range context dependencies for the input data.

\textbf{Transformer-based Methods.}
Several Transformer-based methods for point cloud classification have been proposed recently. To learn global features of point clouds by Transformer, Point Cloud Transformer (PCT) \cite{guo2021pct} adopts the PointNet \cite{qi2017pointnet} architecture where shared MLP layers are replaced with standard Transformer blocks. By utilizing the offset-attention mechanism and neighborhood information embedding, PCT achieves the state-of-the-art performance in point cloud classification. Han et. al \cite{han2021point} proposed another point-wise approach to learn global features. Specifically, they designed a multi-level Transformer to extract global features of target point clouds with different resolutions, followed by concatenating these features and feeding them into a multi-scale Transformer to get the final global features. Instead of extracting global features in a point-wise manner, Point Transformer (PT) \cite{zhao2021point} applies Transformer layers to local neighborhoods of point clouds, and extracts local features hierarchically through transition down modules. Finally, global features can be obtained by a global average pooling operation. However, this method may incur information redundancy, since the Point Transformer block is applied to all input points of each layer. Additionally, as a pure Transformer architecture (without CNNs), it may result in high computation and memory costs due to massive linear transformation layers in Transformers.

\subsection{2D Image Classification.}

Transformer has achieved a significant success in the field of 2D image processing, which is taken as a viable alternative form to the convolutional neural networks in the visual classification task. Vision Transformer (ViT) \cite{dosovitskiy2020image} is the first to introduce a pure Transformer framework into the field of 2D image processing and achieves the better results compared with CNNs on large datasets. ViT divides the image into a series of patches, taken as the input tokens for the network, followed by applying several Transformer blocks for feature learning. Each Transformer block consists of two core stages: Multi-Head Attention and Feed Forward. To leverage local information and reduce computational complexity, Swin Transformer \cite{liu2021swin} proposed a window-based Transformer algorithm, i.e., applying Transformer to fixed-size windows instead of the global image scale. And to build connections with different non-overlapping windows, Swin Transformer introduced a shifted window module. Because of the hierarchical design and cross-window connection, Swin Transformer surpasses the previous state-of-the-art methods in terms of image classification. There also exist several vision Transformer variants \cite{ranftl2021vision, yuan2021tokens, wang2021pyramid, wang2021end} which explore ways to model local features better, such as replacing the predefined positional embedding or constructing connections between multiple tokens. \cite{zhao2020exploring} investigated a series of variants of self-attention and assessed their effectiveness for image processing. By introducing convolution into the Vision Transformer architecture, CvT \cite{wu2021cvt} merges the benefits of Transformers with the benefits of CNNs for image classification task. It achieves superior performance while maintaining computational efficiency. Similar approaches \cite{feng2020point, hui2021pyramid} are also proposed in the 3D fields for point cloud segmentation and place recognition, but nearly none of them explores the effectiveness of different Transformer variants in such framework.

Inspired by CvT \cite{wu2021cvt}, we propose a multi-scale framework to incorporate convolution operations into the Transformer for point cloud classification, which can achieve improvements of classification accuracy and efficiency. Additionally, we also conduct a detailed investigation on different Transformer variants to explore the one that best fits our framework.

\begin{figure*}[htbp]
  \centering
  \includegraphics[width=0.9\linewidth]{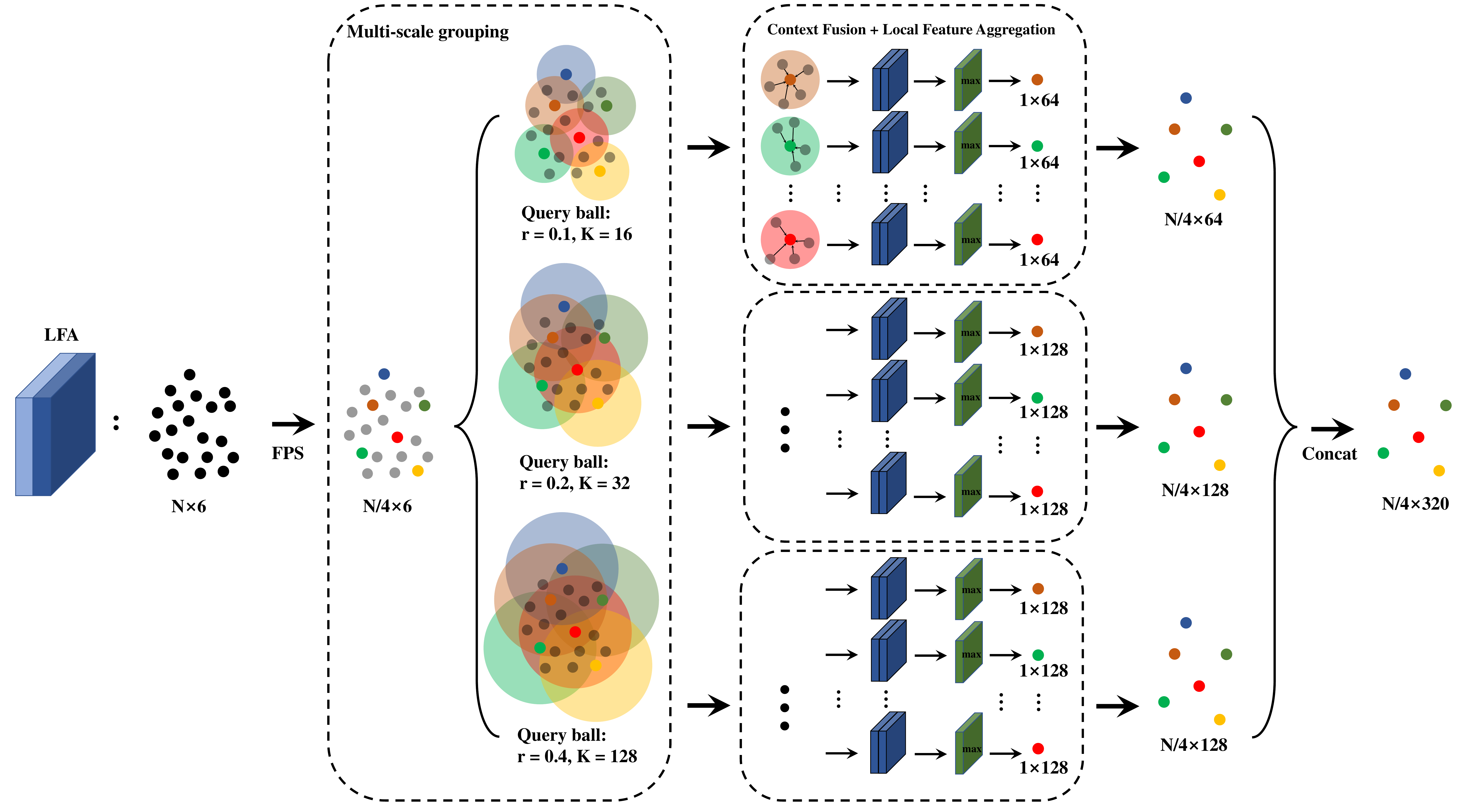}
  \caption{Multi-scale Local Feature Aggregating (LFA) block. Taking module 1 in Figure \ref{fig:overview} as an example, the LFA block has three key steps: Multi-scale grouping, Context Fusion, and Local Feature Aggregation.
  \label{fig:LFA}}
\end{figure*}

\section{3D Convolution-Transformer Network}
\label{sec:method}
In this section, we show how to combine Transformer and convolution in a hierarchical framework for 3D point cloud classification. We begin by presenting the design of our hierarchical network architecture, followed by introducing the convolution-based local feature aggregating and Transformer-based global feature learning process.

\subsection{Overview.}
The overall pipeline of our 3D Convolution-Transformer Network (3DCTN) is shown in Figure \ref{fig:overview}.
We design a hierarchical structure for point cloud classification to improve the efficiency and sensitivity to local geometric layout, which has been proven to be effective by many previous works \cite{liu2021swin, qi2017pointnet++, zhao2021point}.
Taking the original point cloud with normals as the input, the network has two modules operating on downsampling point sets, and each module has two blocks: Local Feature Aggregating (LFA) block and Global Feature Learning (GFL) block, where the former block is based on the graph-convolution, while the later one is based on Transformer. In this way, we effectively combine the strong local modeling ability and high efficiency of CNNs with the remarkable global feature learning ability of Transformers.
The numbers of points in sampling point sets are set to $[N/4, N/16]$ for two modules respectively, where $N$ is the number of input points.
After two modules above, an additional convolution layer with $1 \times 1$ kernel is applied to extend the extracted feature to 1024 dimensions, followed by applying a global max pooling to get the final global feature for the target point cloud. Lastly, an MLP Head layer is utilized to get the global classification logits, which consists of three linear layers with the batch normalization and ReLU.


\subsection{Local Feature Aggregating Block.}
\label{subsec:LFA}
The Local Feature Aggregating (LFA) block is based on Graph Convolution, and it is proposed to achieve local feature extraction in an efficient and effective manner. By aggregating the local features to corresponding center points (sampling points), this block is able to provide efficient discriminative regional feature extraction.

As shown in Figure \ref{fig:LFA}, given the input point cloud, we perform the farthest point sampling (FPS) to obtain the point cloud subset, called the sampling point set. To ensure the diversity of the receptive fields for sampling points, we construct multi-scale neighborhoods of each sampling point by query ball grouping \cite{qi2017pointnet++}. For each neighborhood of a sampling point, we first present a context fusion method to encode and combine the coordinates and feature information of the neighborhood, which has been proven to be effective in \cite{qiu2021pu}, and then adopt Edge Convolution \cite{wang2019dynamic} to aggregate the local features.

\textbf{Context Fusion.}
Given a neighborhood $\chi_{i}$ of a sampling point $x_{i}$, each neighbor point $x_{j}$ has two kinds of contexts: coordinate context $P_{j}$ and feature context $F_{j}$, where the former is used to describe the geometric distribution in 3D space, and the later is used to analyze the semantic information for point cloud classification. To leverage these contexts, we combine both $P_{j}$ and $F_{j}$ as:
\begin{equation}
\mathbb{C}_{j}= concat (F_{j}, P_{j}),
\end{equation}
where $\mathbb{C}_{j}$ is the combined feature of $x_{j}$.
Based on such combined features, we define the relationship between $x_{i}$ and $x_{j}$ as:
\begin{equation}
\Delta \mathbb{C}_{ij}= concat( F_{j} - F_{i}, \mathbb{C}_{i}  ).
\end{equation}
By this way, we are able to encode comprehensive local details for further feature aggregation.

\textbf{Local Feature Aggregation (LFA).}
Having the combined features of points in $\chi_{i}$, we then design a directed graph
$\Psi=\left\{B,E \right\}$ to describe the local structure of $\chi_{i}$, where $B$ represents the neighbor points $\left\{x_{j} \right\}_{j=1}^{j=K}$, $K$ is the number of points in $\chi_{i}$, varying with different modules, and $E$ means edge feature operating on the relationship between $x_{i}$ and $x_{j}$.
The computation of the edge feature $E$ can be defined as:
\begin{equation}
E_{ij}= f(\Delta \mathbb{C}_{ij}),
\end{equation}
where $f(\ast )$ is a nonlinear operator with a set of learnable parameters. There are various ways to choose $f(\ast )$ \cite{wang2019dynamic} , and in our work, we define $f(\ast )$ as:
\begin{equation}
f(\Delta \mathbb{C}_{ij}) = Conv(\Delta\mathbb{C}_{ij}),
\end{equation}
where $Conv$ means a point-wise convolution with $1 \times 1$ kernels.

After computing all edge features, we utilize the maxpooling operation to extract the new feature of $x_{i}$, which expressed as:
\begin{equation}
y_{i} = \operatorname*{\textit{maxpooling}}\limits_{B} \, E_{ij} .
\end{equation}
By this way, we are able to aggregate the local information to the corresponding sampling points for accurate feature representation.

\begin{figure*}[htbp]
  \centering
  \includegraphics[width=0.9 \linewidth]{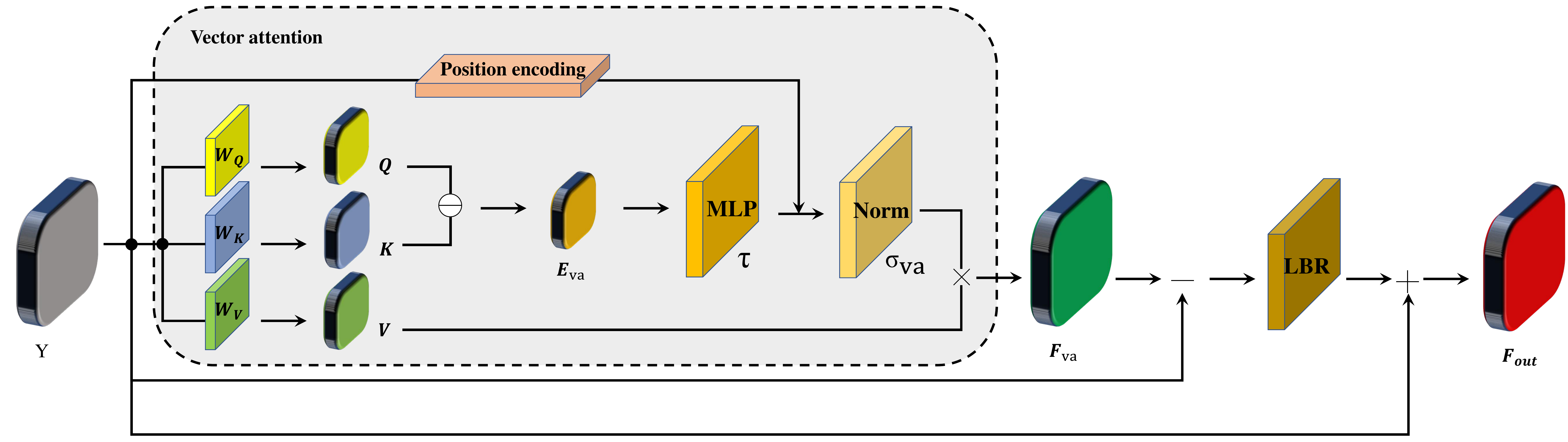}
  \caption{Transformer-based Global Feature Learning (GFL) block. It consists of the offset-attention mechanism and vector attention operator.
  \label{fig:OA}}
\end{figure*}

\subsection{Global Feature Learning Block (GFL).}
\label{subsec:GFL}
Taking the aggregated features $Y=\left\{ y_{i} \right\}_{i=1}^{i=S}$ of sampling points as the input, where $S$ is the number of sampling points, Global Feature Learning (GFL) block  has two main components: self-attention mechanism and position encoding. There is no input (word) embedding in the GFL block, since $Y$ from the LFA block can be considered as the embedded input for the GFL block.
In following subsections, we first elaborate on the self-attention mechanism, \textit{Offset-Attention}, which is proposed by \cite{guo2021pct} and achieves a great improvement for point cloud classification. Next we introduce the self-attention operator, \textit{Vector Attention}, with learnable position encoding.

\textbf{Offset-Attention.}
As show in Figure \ref{fig:OA}, unlike the basic self-attention mechanism, the main idea of offset-attention is to adopt a similar operation as a Laplacian matrix $L = D - E$ \cite{bruna2013spectral} to replace the adjacency matrix E, where $D$ is the diagonal degree matrix. In particular, Offset-Attention (OA) can be defiend as:
\begin{equation}
F_{out} = OA(Y) = LBR(Y - V_A(Y)) + Y,
\end{equation}
where $F_{out}$ is the final output of OA, $V_A(\ast)$ represents the vector attention operator which is detailed in the next subsection, and $Y - V_A(Y)$ is an offset operator  \cite{guo2021pct} analogous to the Laplacian matrix above.
Experiments (Sec. \ref{subsec:investigation}) show that the offset-attention outperforms other self-attention mechanisms.

\textbf{Vector Attention.}
Usually, there are two kinds of self-attention operators: vector attention and scalar attention, where the later has been applied in many previous 3D Transformer works \cite{han2021point, guo2021pct}, while the former has been proven to be more effective than other operators in the fields of 2D image \cite{zhao2020exploring} and 3D point cloud processing \cite{zhao2021point}.

Given input features $Y=\left\{ y_{i} \right\}_{i=1}^{i=S}$, we first compute \textit{Query}, \textit{Key} and \textit{Value} matrices, $Q=\left\{ q_{i} \right\}_{i=1}^{i=S}, K=\left\{ k_{i} \right\}_{i=1}^{i=S}, V=\left\{ v_{i} \right\}_{i=1}^{i=S}$, as:
\begin{equation}
\begin{aligned}
Q =  W_{Q} \times Y ,\\
K =  W_{K} \times Y ,\\
V =  W_{V} \times Y ,\\
\end{aligned}
\end{equation}
where $W_{Q}, W_{K}, W_{V}$ are three learnable weight matrices.
After that, the standard scalar attention can be formulated as:
\begin{equation}
\begin{aligned}
F_{sa} &= E_{sa} \times V \\
&= \sigma_{sa} (Q \times K^{-1} + \rho_{sa} ) \times V,
\end{aligned}
\end{equation}
where $F_{sa}$ is the output feature of the scalar attention, $E_{sa}$ is the adjacency matrix calculated by the scalar product between $Q$ and $K^{-1}$, $\sigma_{sa}$ is a normalization function: \textit{scale} + \textit{softmax}, and $\rho_{sa}$ represents positional encoding. Essentially, $F_{sa}$ is generated by computing the weighted sum of all vectors in $V$, according to the adjacency matrix $E_{sa}$.

Unlike the way of generating the adjacency matrix in the scalar attention, vector attention used in our paper performs a channel-wise subtraction between $Q$ and $K$, which can be described as:
\begin{equation}
\label{eq:vector_attention}
\begin{aligned}
F_{va} &= E_{va} \cdot V \\
&= \sigma_{va} (\tau (Q \ominus  K) + \rho_{va} ) \cdot \bar{V},
\end{aligned}
\end{equation}
where $F_{va}$ is the output feature of the vector attention,
$E_{va}$ is the channel-wise adjacency matrix, $\bar{V}$ is the expanded \textit{Value} matrix to ensure the same shape of two terms in the equation. Specifically, we assume that the shape of $E_{va}$ is $(B, N, N, D)$, where $B$ means the batch size, $D$ means the feature dimension of $q_{i}$. Accordingly, the shape of $V$ is $(B, N, D)$. To ensure the consistent shape, we expand $V$ to $\bar{V}$, which has the same shape, $(B, N, N, D)$, as $E_{va}$. $\tau(\ast)$ represents an MLP operation to produce the attention map, $\sigma_{va}$ is a normalization function: \textit{softmax} + \textit{$l_{1}$ normalization}, $\rho_{va}$ represents positional encoding which is detailed in the next subsection, and $Q \ominus  K$ is defined as:
\begin{equation}
Q \ominus  K =
\begin{bmatrix}
q_{1} - k_{1} & q_{1} - k_{2} & ... & q_{1} - k_{S} \\
q_{2} - k_{1} & q_{2} - k_{2} & ... & q_{2} - k_{S} \\
... & ... & ... & ... \\
q_{S} - k_{1} & q_{S} - k_{2} & ... & q_{S} - k_{S} \\
\end{bmatrix}
\end{equation}

Different from the scalar product, $Q \ominus K$ is designed to measure the difference of corresponding channels between two feature vectors like $q_{m}$ and $k_{n}$. Compared with the scalar attention, vector attention tends to be more flexible and expressive since each channel of the output feature can be modulated according to the channel-wise adjacency matrix.

\textbf{Position encoding.}
In vector attention, we introduce a learnable position encoding to fuse the local spatial information, which is essential for local feature representation. Similar as standard 2D-aware position embedding \cite{dosovitskiy2020image}, we define the 3D position encoding scheme based on the relative coordinates $\mathbb{P} = \left\{ P_{i} \right\}_{i=1}^{i=S} $ of points in $\chi_{i}$, which can be expressed as:
\begin{equation}
\begin{aligned}
\rho_{va} = \xi (\mathbb{P} \ominus \mathbb{P} ),
\end{aligned}
\end{equation}
where $\mathbb{P} \ominus \mathbb{P}$ is a matrix representing 3D relative coordinates of points in $\chi_{i}$:
\begin{equation}
\mathbb{P} \ominus  \mathbb{P} =
\begin{bmatrix}
P_{1} - P_{1} & P_{1} - P_{2} & ... & P_{1} - P_{S} \\
P_{2} - P_{1} & P_{2} - P_{2} & ... & P_{2} - P_{S} \\
... & ... & ... & ... \\
P_{S} - P_{1} & P_{S} - P_{2} & ... & P_{S} - P_{S} \\
\end{bmatrix}
\end{equation}
and $\xi$ represents an MLP operation which consists of two linear layers separated by batch normalization and ReLU, and it is used to extend the feature dimension of relative coordinates from $3$ to the same dimension as $Q$ and $K$, to achieve channel-wise summation in Eq.\eqref{eq:vector_attention}.

\section{Experiments}
\label{sec:Experiments}

In this section, we first introduce the implementation of our algorithm, including hardware configuration and hyperparameter settings. Secondly, we evaluate the performance of our network on the public dataset, ModelNet40 \cite{wu20153d}, and compare it with the state-of-the-art works in point cloud classification in terms of accuracy and processing efficiency. Thirdly, we perform a series of ablation studies to verify the effectiveness of each main component in our framework. Fourthly, we conduct a detailed investigation and analysis on a series of Transformer variants (Figure \ref{fig:investigation}) in our network for better performance. Lastly, we illustrate the interpretability of our network by heat map visualization results, which shows that our method is able to understand different kinds of point clouds by their distinctive features.

\subsection{Implementation Details.}
We implement the classification network with Pytorch and train it on a NVIDIA Tesla V100 GPU.
The network is trained with the SGD Optimizer, with a momentum of $0.9$ and weight decay of 0.0001. The initial learning rate is set to $0.01$, with a cosine annealing
schedule to adjust the learning rate at every epoch. We train the network for 250 epochs and set the batch size as 16. All the involved parameters of our method are empirically set for better performance.

\subsection{Comparison to the State of the Art.}
We compare our method with the state-of-the-art works including Transformer-based methods and other deep learning-based methods, in terms of classification accuracy and efficiency.

\begin{figure*}[htbp]
  \centering
  \includegraphics[width=0.9\linewidth]{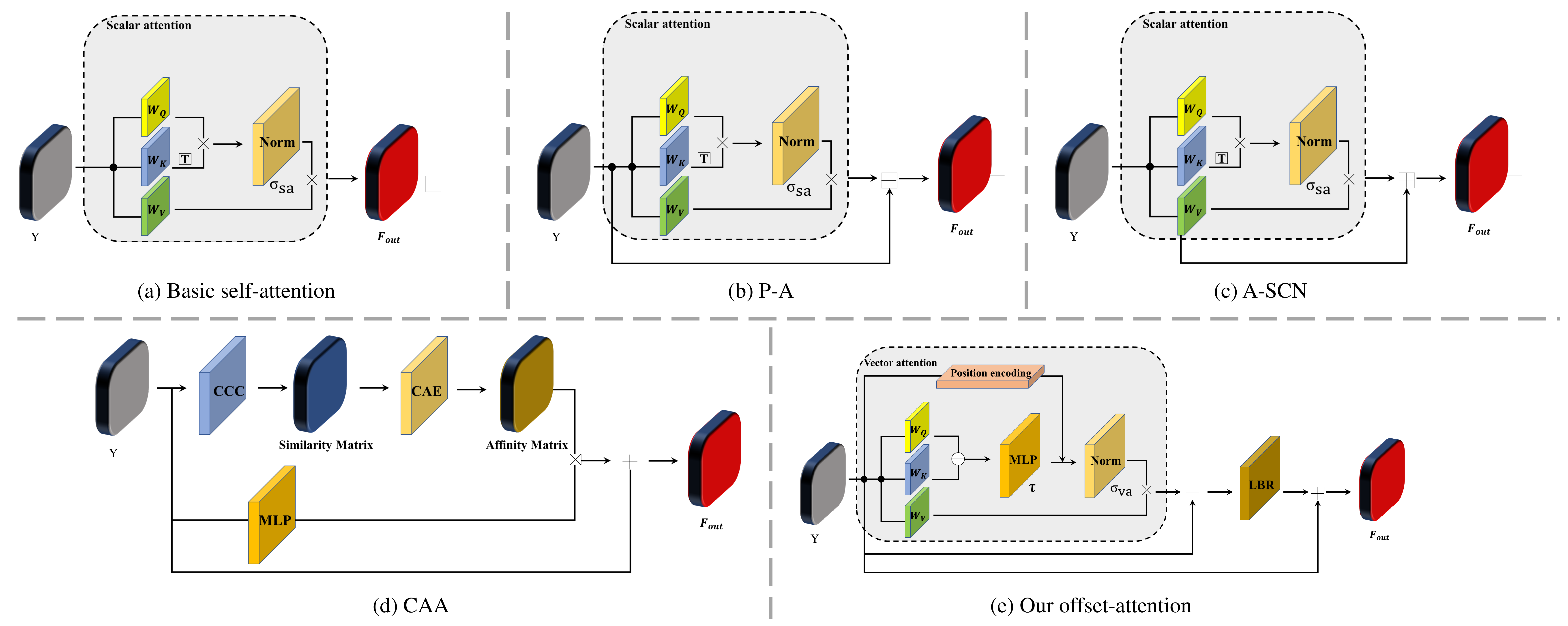}
  \caption{Architectures of various self-attention mechanisms for 3D point cloud processing. (a) Basic self-attention. (b) P-A \cite{feng2020point}. (c) A-SCN \cite{xie2018attentional}. (d) CAA \cite{qiu2021geometric}. (e) Our offset-attention.
  \label{fig:investigation}}
\end{figure*}

\textbf{Dataset and Metrics.}
The ModelNet40 \cite{wu20153d} dataset is widely used in 3D point cloud classification. It consists of $12311$ CAD-like models in 40 object categories, which have been split into $9843$ training models and $2468$ testing models. Each model is downsampled uniformly to 1024 points with normals, as the input of the network, following PointNet \cite{qi2017pointnet}.
For evaluation metrics, we utilize the mean accuracy operated on each category ($mAcc$) and the overall accuracy ($OA$) operated on all classes, which are formulated as:
\begin{equation}
\begin{aligned}
mAcc &= \frac{\sum_{i=1}^{K}\frac{T_{i}}{N_{i}}}{K},\\
OA &= \frac{T}{N},
\end{aligned}
\end{equation}
where $T$ is the number of all correctly predicted point clouds, $T = \sum_{i=1}^{K}T_{i}$, $T_{i}$ is the number of correctly predicted point clouds in class $i$, $K$ is the number of classes in the dataset, $N$ is the number of all point clouds in the dataset, $N = \sum_{i=1}^{K}N_{i}$ and $N_{i}$ is the number of point clouds in class $i$.
Additionally, we adopt total parameters and FLOPs of the involved networks to evaluate the efficiency.

\textbf{Performance Comparison.}
As shown in Table \ref{tab:acc_comparison}, compared to Transformer-based approaches, our method achieves a higher value of $mAcc$ and a competitive value of $OA$, thanks to the combination of CNNs and Transformers. Specifically, our method obtains a Top-1 $mAcc$, 91.2$\%$, which is $0.6\%$ higher than PointTransformer \cite{zhao2021point} and $0.2\%$ higher than CAA \cite{qiu2021geometric}, which means that our method has robust classification performance for different types of point clouds. Compared to other deep learning-based approaches, our method still achieves the highest $mAcc$ thanks to the strong global feature learning ability of Transformers, and outperforms most of state-of-the-art approaches like Graph Convolution-based method DGCNN \cite{wang2019dynamic}, point-based convolution method PointCNN \cite{li2018pointcnn}, and attention-based method Point2Sequence \cite{liu2019point2sequence} in terms of $OA$.

\begin{table}[h]
 \centering

 \caption{Comparison with state-of-the-art methods on ModelNet40 in terms of classification accuracy.
 }
 \label{tab:acc_comparison}
 \begin{tabular}{c|cc}
  \hline
   {Models} & mAcc  & OA    \\
 \hline

  {Other Learning-based Methods} \\
   \hline
  {3DShapeNets}\cite{wu20153d} & 77.3\%  & 84.7\%     \\
  {VoxNet}\cite{maturana2015voxnet}    & 83.0\% & 85.9\%     \\
  {Subvolume}\cite{qi2016volumetric}  & 86.0\%  & 89.2\%     \\
  {PointNet}\cite{qi2017pointnet} & 86.0\%  & 89.2\%    \\
  {PointNet++}\cite{qi2017pointnet++} & -  & 91.9\%     \\
  {diffConv}\cite{lin2021diffconv} & 90.4\%  & 93.2\%    \\
  {PointWeb}\cite{zhao2019pointweb} & 89.4\%  & 92.3\%     \\
  {CurveNet}\cite{muzahid2020curvenet} & 90.4\%  & 93.1\%     \\
  {PointCNN}\cite{li2018pointcnn} & 88.1\%  & 92.2\%     \\
  {Point2Sequence}\cite{liu2019point2sequence} & 90.4\%  & 92.6\%     \\
  {DGCNN}\cite{wang2019dynamic} & 90.2\%  & 92.2\%     \\
  {FatNet}\cite{kaul2021fatnet} & 90.6\%  & 93.2\%     \\
  \hline
  {Transformer-based Methods} \\
   \hline
  {A-SCN}\cite{xie2018attentional} & 87.6\%  & 90.0\%     \\
  {PATs}\cite{yang2019modeling} & -  & 91.7\%     \\
  {GAPNet}\cite{chen2019gapnet} & 89.7\%  & 92.4\%     \\
  {LFT-Net}\cite{gao2022lft} &89.7\% &93.2\%  \\
  {3DETR}\cite{misra2021end} &89.9\% &91.9\%  \\
  {PointTransformer}\cite{engel2021point} &- &92.8\%  \\
  {MLMST}\cite{han2021point} & -  & 92.9\%     \\
  {PointCloudTransformer}\cite{guo2021pct} & -  & 93.2\%     \\
  {PointTransformer}\cite{zhao2021point} & 90.6\%  & 93.7\%     \\
  {CloudTransformers}\cite{mazur2021cloud} & 90.8\%  & 93.1\%     \\
  {CAA}\cite{qiu2021geometric} & 91.0\%  & \textbf{93.8\%}     \\
   \hline
  {Ours (Single-scale)}    & \textbf{91.2\%}  & 92.7\%     \\
  {Ours (Multi-scale)}    & \textbf{91.2\%}  & 93.3\%     \\

  \hline
 \end{tabular}
\end{table}

Additionally, thanks to our hierarchical structure and efficient local CNN blocks, our method with single-scale neighborhood has fewer parameters and FLOPs than other Transformer-based approaches. As shown in Table \ref{tab:eff_comparison}, our single-scale model produces a reduction of $80.1\%$ parameters and $93.7\%$ FLOPs, compared with PointTransformer.

\begin{table*}[htbp]
 \centering

 \caption{Comparison with state-of-the-art methods on ModelNet40 in terms of classification efficiency.
 }
 \label{tab:eff_comparison}
 \begin{tabular}{c|c|cccc}
  \hline
    \multicolumn{2}{c}{Models} & Parameters (MB)  & FLOPs (GB) & mAcc  & OA    \\
 \hline

  \multirow{6}{*}{Other Learning-based Methods}&{PointNet}\cite{qi2017pointnet} & 3.47  & 0.45  & 86.0\%  & 89.2\%     \\
  &{PointNet++ (SSG)}\cite{qi2017pointnet++}  & \textbf{1.48}  & 1.68  & -  & 91.9\%     \\
  &{PointNet++ (MSG)}\cite{qi2017pointnet++}  & 1.74  & 4.09  & -  & 91.3\%     \\
  &{diffConv}\cite{lin2021diffconv}  & 2.08  & \textbf{0.16}  & 90.4\%  & 93.2\%    \\
  &{CurveNet}\cite{muzahid2020curvenet}  & 2.14  & 0.30  & 90.4\%  & 93.1\%     \\
  &{DGCNN}\cite{wang2019dynamic}  & 1.81  & 2.43  & 90.2\%  & 92.2\%     \\
  \hline
  \multirow{3}{*}{Transformer-based Methods}
  &{GAPNet}\cite{chen2019gapnet} &1.90 &-& 89.7\%  & 92.4\%     \\
  &{PointTransformer}\cite{engel2021point}  & 21.0  & 5.05  & -  & 92.8\%     \\
  &{PointCloudTransformer}\cite{guo2021pct}  & 2.88  & 2.32  & -  & 93.2\%     \\
  &{PointTransformer}\cite{zhao2021point}  & 9.14  & 17.1  & 90.6\%  & \textbf{93.7\%}     \\
   \hline
  \multicolumn{2}{c|}{Ours (Single-scale)}  & 1.82  & 1.07  & \textbf{91.2\%}  & 92.7\%     \\
  \multicolumn{2}{c|}{Ours (Multi-scale)}  & 4.22  & 4.05  & \textbf{91.2\%}  & 93.3\%     \\
  \hline
 \end{tabular}
\end{table*}

\begin{table*}[htbp]
 \centering
  \renewcommand{\arraystretch}{1.5} 
 \caption{Ablation study.
 }
 \label{tab:ablation}
 \begin{tabular}{c|c|c|c|c|c|c}
  \hline
    \multicolumn{2}{c|}{Ablation}  & mAcc  & OA  & Parameters (MB)  & FLOPs (GB) & TE (s) \\
 \hline

  {Hierarchical structure} & \XSolidBrush & -  & - & 4.22  & 92.36 & -      \\
   \hline

  {Multi-scale neighborhood}  &\XSolidBrush & 91.2\%  & 92.7\% & 1.82  & 1.08 &  \textbf{186}  \\
     \hline

  \multirow{2}{*}{Local feature aggregating}  &\XSolidBrush & 87.5\%  & 90.6\% & 4.19  &  \textbf{0.33 }  & 266 \\
  \cline{2-7}
      &Transformer-based & -  & - & 4.99  & 51.54  & - \\
     \hline

  \multirow{2}{*}{Global feature learning}  &\XSolidBrush & 91.0\%  & 93.1\% & \textbf{1.76  }& 2.40  &289   \\
  \cline{2-7}
     &Graph Convolution-based & 90.4\%  & 93.2\% & 3.26  & 26.37  & 525   \\
     \hline

  {Position encoding}  &\XSolidBrush & 90.9\%  & 93.1\% & 4.22  & 4.06  &298    \\
     \hline

  {Self-attention mechanism}  &Basic mechanism & 90.4\%  & 91.9\% & 4.22  & 4.06  & 297   \\

    \hline

  {Self-attention operator}  &Scalar attention & 90.5\%  & 93.1\% & 4.21  & 2.65 & 255 \\
  \hline
  \hline
  \multicolumn{2}{c|}{3DCTN}  & \textbf{91.2\%}  & \textbf{93.3\%} & 4.22  & 4.06  &303 \\
  \hline
 \end{tabular}
\end{table*}

\begin{figure*}[htbp]
  \centering
  \includegraphics[width=0.8\linewidth]{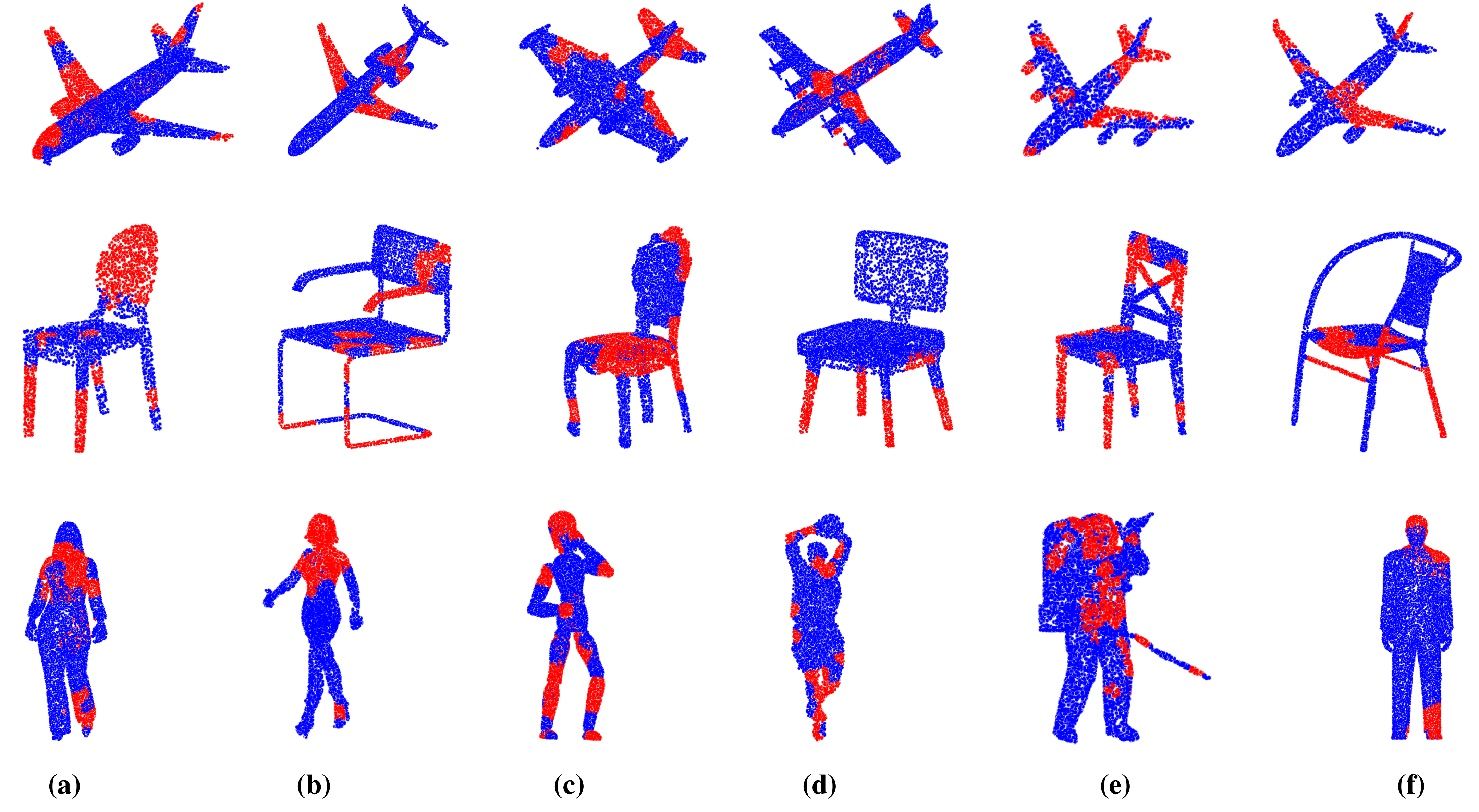}
  \caption{Heat map visualization.
  \label{fig:heatmap}}
\end{figure*}

\subsection{Ablation Study.}
In this section, we perform various ablation experiments to evaluate the effectiveness of each main component of our framework on ModelNet40.

\textbf{Hierarchical Structure.}
The proposed hierarchical structure can aggregate local features effectively, with the significant reduction of computational and memory costs. As shown in Table \ref{tab:ablation} (row 2), for the network without the hierarchical structure, it is inevitable to require more computational and memory costs, since the multi-scale local feature aggregating and global feature learning are performed on each point. This study suggests that the proposed hierarchical structure is effective to improve the efficiency for our framework.

\textbf{Multi-scale Strategy.}
We compare the multi-scale local feature aggregating with the single-scale (middle scale) way , and show the results in Table \ref{tab:ablation} (row 3). On one hand, in terms of classification accuracy, the performance with single-scale strategy is lower than multiple-scale strategy, which demonstrates that multiple receptive fields are benefit to the local feature aggregating. On the other hand, in terms of efficiency, single-scale strategy leads to a significant reduction, 57$\%$ parameters, 73$\%$ FLOPs, and $37\%$ TE (Time per Epoch).

\textbf{Local Feature Aggregating.}
 We study the effectiveness of the proposed LFA block (Sec. \ref{subsec:LFA}), and Table \ref{tab:ablation} (row 4) shows the results. We first remove the LFA block, making the framework a pure Transformer network. From the results, without the local feature aggregating, the performance drops significantly, which indicates the importance of the LFA block. Next we replace Graph Convolution in LFA with Transformer, and we can see that the total parameters and FLOPs of the network are much higher than the original. These results validate local feature aggregating based on Graph Convolution has higher efficiency and better local modeling ability.

\textbf{Global Feature Learning.}
We also study the choice of different global feature learning methods in the GFL block (Sec. \ref{subsec:GFL}), and the results are shown in Table \ref{tab:ablation} (row 5). From the results, the removal of the GFL block causes a drop in classification performance, and when we replace Transformer with Graph Convolution, the performance drops again, which indicates the superiority of the Transformer-based GFL block.

\textbf{Position Encoding.}
The position encoding in Eq.\eqref{eq:vector_attention} can introduce the spatial difference of the input words (points) to the attention map. We conduct an ablation study on that and show results in Table \ref{tab:ablation} (row 6). Without the position encoding, both $mAcc$ and $OA$ are lower than the original, indicating the effectiveness of such component.

\textbf{Self-attention Mechanism.}
We compare the offset-attention in the GFL block with the basic self-attention mechanism (shown in Figure \ref{fig:investigation}(a)) in Table \ref{tab:ablation} (row 7). From the results, by using the basic self-attention mechanism, we observe a $0.8\%$ and $1.4\%$ drop in $mAcc$ and $OA$ respectively. This suggests that the offset-attention outperforms the basic self-attention mechanism. Additionally, we conduct a detailed investigation about different self-attention mechanisms for our framework, and the architectures of there mechanisms are shown in Fig \ref{fig:investigation}, please see Sec. \ref{subsec:investigation} for more results.

\textbf{Self-attention Operator.}
We also investigate the types of the self-attention operator used in the GFL block. Generally, there are two commonly-used self-attention operators: vector attention and scalar attention. As shown in Table \ref{tab:ablation} (row 8), scalar attention leads to a drop of accuracy ($0.7 \%$ in $mAcc$ and  $0.2\%$ in $OA$), compared with vector attention. This demonstrates the superiority of vector attention. And Sec. \ref{subsec:investigation} shows a detailed investigation about different variants of the self-attention.

\begin{table*}[t]
 \centering
\renewcommand{\arraystretch}{1.5} 
 \caption{Investigation of self-attention mechanisms.
 }
 \label{tab:sa_mechanism}
 \begin{tabular}{c|ccccc}
  \hline
    {Self-attention mechanisms} & mAcc  & OA & Parameters (MB)  & FLOPs (GB)  &TE(s)    \\
 \hline

  {Basic} & 90.4\%  & 91.9\%  & 4.22  & 4.06  &297    \\
  {A-SCN}\cite{xie2018attentional}  & 90.9\%  & 92.9\%  & 3.72  & 2.60  & 277    \\
  {P-A}\cite{feng2020point}  & 90.8\%  & 92.8\%  & 4.21  & 2.65  &278    \\
  {CAA}\cite{qiu2021geometric}  & 91.1\%  & 93.2\%  & \textbf{2.27}  & \textbf{2.45}  &\textbf{277}   \\
  {Offset-attention}\cite{guo2021pct}  & \textbf{91.2\%}  & \textbf{93.3\%}  & 4.22  & 4.06 &303     \\
  \hline
 \end{tabular}
\end{table*}

\begin{table*}[t]
 \centering
\renewcommand{\arraystretch}{1.5} 
 \caption{Investigation of self-attention operators.
 }
 \label{tab:sa_operator}
 \begin{tabular}{c|c|ccccc}
  \hline
  \multicolumn{2}{c}{Self-attention operators} & mAcc  & OA  & Parameters (MB)  & FLOPs (GB) &TE(s) \\
 \hline

  \multirow{1}{*}{Scalar attention }&{Dot product} & 90.5\%  & 93.1\%  & 4.21  & 2.65 &255     \\
  \hline
  \multirow{5}{*}{Vector attention}&{Concatenation} & 89.6\%  & 93.1\%  & 4.43  & 6.17  &331  \\
  &{Summation} & 90.6\%  & 92.9\%  & 4.22  & 4.06 &294  \\
  &{Subtraction} & \textbf{91.2\%}  & \textbf{93.3\%}  & 4.22  & 4.06 &303   \\
  &{Division} & 88.8\%  & 91.9\%  & 4.22  & 4.06  &300   \\
  &{Hadamard product} & 90.1\%  & 92.9\%  & \textbf{4.22}  & \textbf{4.06}   &\textbf{289}\\

  \hline
 \end{tabular}
\end{table*}

\subsection{Investigation on 3D Attentions.}
\label{subsec:investigation}
In this section, we conduct a detailed investigation on self-attention mechanisms and operators for better performance in point cloud classification. This investigation is also expected to provide some benefit references for Transformer-based classification works.

\textbf{Self-attention Mechanisms.}
We collect a series of self-attention mechanisms widely used in the 3D point cloud processing. As shown in Figure \ref{fig:investigation}, the basic self-attention mechanism includes three linear layers to generate \textit{Query}, \textit{Key} and \textit{Value} matrices. The attention map is estimated by comparing \textit{query} and \textit{key}, and then normalized to limit the variance of the matrix. The final output can be obtained by multiplying the attention matrix and \textit{value} matrix. Attentional ShapeContextNet (A-SCN) \cite{xie2018attentional} and Point-Attention (P-A) \cite{feng2020point} have similar architectures to the basic self-attention mechanism, while the only difference is that both of them apply residual connection operations to strengthen the connection between the input and output.
Therefore, they both achieve better classification results than the basic self-attention mechanism. Different from above point-wise mechanisms, Channel-wise Affinity Attention (CAA) \cite{qiu2021geometric} focuses on the channel space, and achieves an outstanding performance in point cloud classification. It first presents a Compact Channel-wise Comparator block (CCC) to generate the similarity matrix efficiently, followed by introducing a Channel Affinity Estimator block (CAE) to generate the affinity matrix which is able to sharpen the attention weights and reduce the redundant information. Lastly, the output is calculated by multiplying the affinity matrix and \textit{value} matrix, with the same residual connection as P-A \cite{feng2020point}. From results in Table \ref{tab:sa_mechanism}, the offset-attention mechanism achieves a better result than the other, which suggests that it is more suitable to our framework. Additionally, CAA\cite{qiu2021geometric} also gets satisfactory performance both in accuracy and efficiency.

\textbf{Self-attention Operators.}
The effectiveness of different self-attention operators has been studies in the field of 2D image processing by \cite{zhao2020exploring}, but there is still no such investigation in point cloud classification. Self-attention operators can be generally divided into two types: scalar attention and vector attention.
The former operates on feature-level similarity estimation, while the later operates on channel-level estimation. The detailed definition of each operator is shown below:
\begin{equation}
\begin{aligned}
\textbf{scalar attention}: F_{sa} &= E_{sa} \times V = \sigma_{sa} (Q \times K^{-1}) \times V, \\
\textbf{vector attention}:  F_{va} &= E_{va} \cdot V = \sigma_{va} (\tau (\Delta  (Q, K)) \cdot \bar{V}, \\
\end{aligned}
\end{equation}
where the representative form of scalar attention is the \textit{dot product}, and $\Delta $ in the second equation represents various forms of vector attention, which can be formulated as:
\begin{equation}
\Delta  = \begin{bmatrix}
\delta(q_{1}, k_{1})  & \delta(q_{1}, k_{2}) & ... & \delta(q_{1}, k_{S}) \\
\delta(q_{2}, k_{1})  & \delta(q_{2}, k_{2}) & ... & \delta(q_{2}, k_{S}) \\
... & ... & ... & ... \\
\delta(q_{S}, k_{1})  & \delta(q_{S}, k_{2}) & ... & \delta(q_{S}, k_{S}) \\
\end{bmatrix}
\end{equation}
where $\delta$ can be expressed as different forms:
\begin{equation}
\begin{aligned}
&\textit{Concatenation} : \delta(q_{i}, k_{i}) = \left [ q_{i}, k_{i} \right ], \\
&\textit{Summation} : \delta(q_{i}, k_{i}) = q_{i} + k_{i}, \\
&\textit{Subtraction} : \delta(q_{i}, k_{i}) = q_{i} - k_{i}, \\
&\textit{Division}  : \delta(q_{i}, k_{i}) = q_{i} /  k_{i}, \\
&\textit{Hadamard product} : \delta(q_{i}, k_{i}) = q_{i} \cdot  k_{i}. \\
\end{aligned}
\end{equation}
We apply all operators above to our framework, to evaluate the their performance. From the results in Table \ref{tab:sa_operator}, we can see that the scalar attention outperforms most channel-wise operators, while the subtraction-form vector attention achieves higher accuracy than it, which suggests that the subtraction-form vector attention operator is more suitable for our framework in point cloud classification.

\subsection{Heat Map visualization.}
To highlight the interpretability of our network, we generate attention map visualization results by utilizing the Grad-CAM method \cite{zhou2016learning} which is commonly used in the 2D fields. As shown in Figure \ref{fig:heatmap}, heat maps show the different regions of interest (ROI) of the network for different types of point clouds. For the airplane category, our network focuses more on the wing and tail parts of the airplane, which are also the key geometric features different from other categories. For the chair category, the ROI of our network are the chair leg and back, which make them distinguishable from desks. And for the person category, our network focuses more on the head and limbs, which is also consistent with our common sense of person classification. Specially, as shown in Figure \ref{fig:heatmap} row 3 (e), when there are some interference point sets (like backpack) in the person point cloud, our network are still able to focus on the key areas of the person, which illustrates the robust performance of our network.

\section{Conclusion}
\label{sec:conclusion}
In this paper, we propose a hierarchical framework that incorporates convolution into Transformer for 3D point cloud classification. Taking a point cloud with normals as the input, our method has two main modules to extract the global feature progressively. In each module, we first aggregate the multi-scale local features by Graph Convolution and then learn the global features by Transformer, which is able to combine the strong local modeling ability and high efficiency of CNNs with the remarkable global feature learning ability of Transformers. After that, we add an additional convolution layer with a $1 \times 1$ kernel, a maxpooling operation, and an MLP head layer sequentially, to generate final classification results. 
To explore the better classification performance, we conduct a detailed investigation on a series of variants of Transformer. 
To demonstrate the effectiveness of the proposed method, we design a number of ablation experiments on main components of our framework. 
Extensive experiments on ModelNet40 dataset prove the effectiveness of our method, and show that it achieves state-of-the-art classification performance in terms of both accuracy and efficiency.

\textbf{Future work.} There have been many studies on the combination of the convolution and Transformer in 2D images, and we believe that it will also be a promising research direction in 3D point cloud processing. In the future, we will further develop our framework and extend it to more complex 3D applications, such as point cloud segmentation and object detection.

\bibliographystyle{IEEEtran}
\bibliography{mybibfile}



\vfill

\end{document}